\pdfoutput=1
\documentclass{article} % For LaTeX2e
\usepackage[final]{colm2026_conference}

\usepackage{microtype}
\usepackage{hyperref}
\usepackage{url}
\usepackage{booktabs}
\usepackage{amsmath}
\usepackage{graphicx}
\usepackage{booktabs}
\usepackage{subcaption}
\usepackage{float}
\usepackage{wrapfig}
\usepackage{lineno}

\definecolor{darkblue}{rgb}{0, 0, 0.5}
\hypersetup{colorlinks=true, citecolor=darkblue, linkcolor=darkblue, urlcolor=darkblue}

\title{Token-Level Off-Policy Learning for Faithful Generation\\Under Distribution Shift}

\newcommand{\aspace}{\hspace{2em}}
\author{Zitong Huang$^\star$ \aspace Gustavo Lucas Carvalho$^\star$ \aspace Deqing Fu \aspace Robin Jia\\
University of Southern California\\
\texttt{\{chuang95,lucasdec,deqingfu,robinjia\}@usc.edu}
}

\usepackage{color-edits}
\addauthor[Cynthia]{ch}{red}
\addauthor[Gustavo]{gc}{green}
\addauthor[Deqing]{df}{magenta}

\begin{document}

\ifcolmsubmission
\linenumbers
\fi

\maketitle

\begin{abstract}
We propose \textbf{T}oken-Level \textbf{O}ff-\textbf{P}olicy \textbf{L}abeling (TOPL), an off-policy training paradigm that reframes post-training as a token-level correctness prediction task. 
Our key intuition is that by training the model to distinguish good and bad tokens in a response, we naturally guide the model towards generating good tokens, while avoiding the pitfalls that come with directly training the model to generate off-policy tokens.
Experiments on document summarization tasks show that TOPL achieves strong out-of-distribution generalization across 11 datasets against a diverse set of sequence-level and token-level baselines. We further demonstrate that TOPL transfers effectively to machine translation, suggesting that its benefits generalize across different faithful generation tasks.
Through ablation studies, we confirm that our token-level learning signal is critical to good performance; sequence-level analogues do not confer similar benefits. 
Finally, we show that TOPL induces interpretable model updates: the LoRA adapters learned through TOPL function as linear classification heads and steering vectors.
{\renewcommand{\thefootnote}{$\star$}\footnotetext[1]{Equal Contribution}}
\end{abstract}

\section{Introduction}
% FINAL VERSION

% While LLMs have achieved strong performance across a wide range of natural language understanding and generation tasks, they still struggle to generalize under distribution shifts 
While LLMs have achieved strong performance across a wide range of natural language understanding and generation tasks, they still struggle with hallucinations and factuality~\citep{bang2025hallulens, huang2025survey, ji2023survey}. To mitigate said bad model behaviors, which often arise from reliance on memorized patterns, current approaches rely on reinforcement learning (RL)-based methods, including both on-policy and off-policy variants. On-policy methods, such as Group Relative Policy Optimization (GRPO; ~\citealp{shao2024deepseekmath}) and its variants~\citep{yu2025dapo, liu2025understanding}, have shown promising improvements in generalization, but suffer from substantial complexity due to online sampling and repeated rollouts. Off-policy methods, such as Direct Preference Optimization (DPO; ~\citealp{rafailov2023direct}), provide a simpler alternative by reusing preference data to directly train policy models. However, their ability to generalize under distribution shift remains limited~\citep{xu2024dpo, ma2025gradient}.

To further improve factuality while retaining the simplicity of off-policy methods, we propose \textbf{T}oken-Level \textbf{O}ff-\textbf{P}olicy \textbf{L}abeling (TOPL), an off-policy training paradigm that reframes post-training as a token-level correctness prediction task. Instead of directly optimizing next-token prediction, TOPL learns token-level correctness features through a binary classification objective trained using Low-Rank Adaptation (LoRA).
% Similar in spirit to DPO, which repurposed reward modeling data with a custom loss function, our method effectively trains a reward model that, with slight modifications, can be used as a generation model without any extra training. 

We primarily evaluate TOPL on document summarization tasks, where factual consistency remains challenging in OOD settings. We utilize a synthetic training dataset based on FAVA~\citep{mishra2024fine}, which introduces token-level perturbations for training. Compared with supervised fine-tuning (SFT), DPO, and existing token-level baselines such as Token-Level DPO~\citep{zeng2024token}, Token-Level Detective Reward (TLDR; ~\citealp{fu2025tldr}), and Token-Level Unlikelihood Training~\citep{welleck2019neural}, TOPL achieves strong out-of-distribution performance, attaining the strongest overall results on average across 11 datasets under distribution shift. 
% TOPL achieves strong performance on both in-distribution and out-of-distribution settings, with the strongest overall performance on average across 11 out-of-distribution datasets. 
Further experiments show that sequence-level analogues do not provide similar benefits, highlighting the importance of fine-grained token-level training signals. We also demonstrate that TOPL transfers effectively to machine translation, suggesting that its benefits extend across different faithful generation tasks.

% Compared with supervised fine-tuning (SFT) and sequence-level baselines such as DPO, TOPL achieves strong performance on both in-distribution and out-of-distribution settings, with especially pronounced improvements under distribution shifts.

 % In this view, 
 To understand how a binary classification objective can improve generation, we analyze how LoRA behaves in TOPL through the lens of conditional steering~\citep{lee2025programming}. By re-purposing LoRA-$A$ as a classifier of factual and nonfactual tokens, we demonstrate experimentally how the the LoRA-$A$ components acts as a conditional mechanism that extracts token-level correctness signals. In turn, we show that the signal extracted by LoRA-$A$ is then used as a weight for its corresponding LoRA-$B$ vectors, which function as steering vectors, pushing the hidden representation towards or away from factual behavior. We empirically validate this hypothesis by incrementally changing the proportion of the LoRA-$B$ vectors added to the hidden state, showing that increasing LoRA-$B$'s contribution improves the model's factuality.

Overall, our contributions are as follows: (1) we introduce TOPL, a novel off-policy method that reframes post-training as token-level correctness classification, achieving strong out-of-distribution performance relative to both token-level and sequence-level baselines across text generation tasks.
 And (2) we provide an analysis of why TOPL is effective, offering a hypothesis grounded in its connection to conditional steering vectors.

\begin{figure}[tp]
    \centering
    \includegraphics[width=\linewidth]{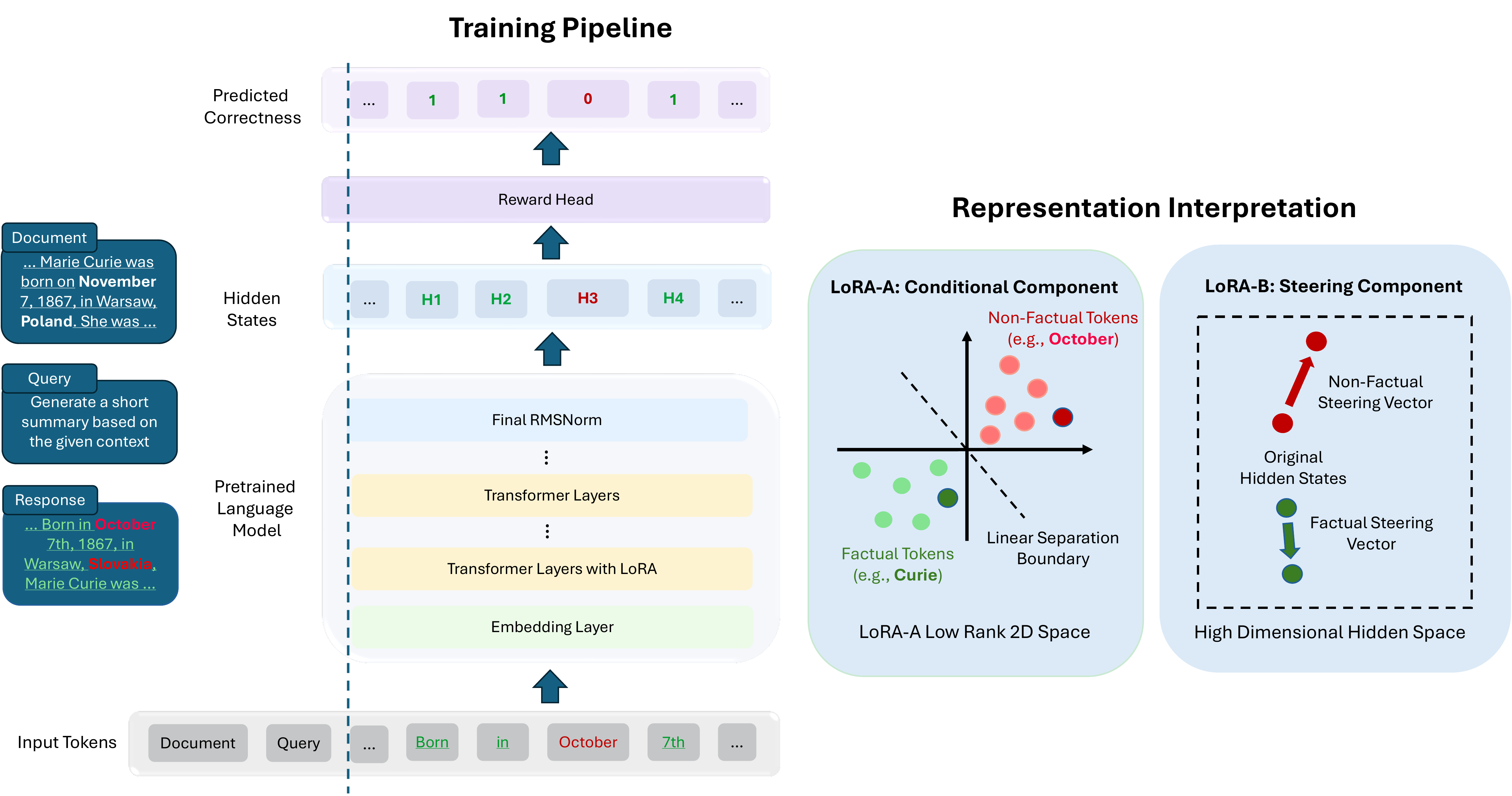}
    \caption{\textbf{Overview of TOPL training and its representation-space interpretation.}
    \textbf{Left: Training pipeline.} Given a document, a query, and a response containing both factual and non-factual tokens, the inputs are processed by a large language model (LLM) with LoRA adapters to obtain token-level hidden representations. Instead of performing standard next-token prediction with a language modeling head, TOPL replaces the objective with a binary reward head that predicts whether each token is {\color{teal} factual} or {\color{red} non-factual}. 
    \textbf{Right: Representation-space interpretation.} Inside the hidden representation space of the LLM, LoRA-$A$ acts as a conditional projection mechanism that separates factual and non-factual token representations within a low-rank subspace, while LoRA-$B$ defines steering directions that shift hidden states toward different behavioral regions. The illustrated example corresponds to a rank-2 LoRA decomposition, where each rank component can be interpreted as an independent conditional steering direction.
    }
    \label{fig:TOPL-model}
\end{figure}

\section{Related Work}

\paragraph{Reinforcement Learning and Preference Optimization.} 

Reinforcement learning from human feedback (RLHF; ~\citealp{christiano2017deep, ziegler2019fine}) is a standard approach for post-training language models with preference or reward signals, including both on-policy and off-policy methods. On-policy methods~\citep{christiano2017deep, schulman2017proximal, yu2025dapo} have shown strong performance in improving generalization, but rely on online sampling and repeated rollouts, introducing substantial complexity during training. Off-policy methods~\citep{meng2023off}, such as Direct Preference Optimization (DPO; ~\citealp{rafailov2023direct}), provide a more lightweight alternative and introduce the perspective that the policy itself can serve as a reward model. Beyond DPO, recent work such as LLaVA-Critic-R1~\citep{wang2025llava} explores unifying critic and policy models by training generative models on preference-labeled data with reinforcement learning. However, these approaches operate at the sequence level, limiting the granularity of supervision. Token-Level Detective Reward (TLDR; ~\citealp{fu2025tldr}) introduces a token-level reward model by assigning labels to individual tokens, and observes that it can also improve generation. However, it still treats the reward model as a separate component, without a unified formulation for directly optimizing generation. In contrast, we propose a method that directly optimizes token-level reward signals as the post-training objective, resulting in a unified single-model formulation that is both fine-grained at the token level and more interpretable.

\paragraph{Model Steering.} 
Model steering methods aim to control the behavior of language models at inference time by modifying their internal representations~\citep{rimsky2024steering, marks2023geometry}. A common approach is representation-based steering, where specific directions in the model’s activation space are identified and used to shift model outputs toward desired behaviors~\citep{li2023inference, turner2023steering, beaglehole2026toward}. A particularly relevant formulation is conditional steering~\citep{lee2025programming}, where the steering direction depends on the input context, enabling more adaptive and fine-grained control over generation. Rather than applying a fixed global modification, conditional steering allows the model to dynamically adjust its behavior based on the underlying representation. In this work, we show that when combined with Low-Rank Adaptation (LoRA), our method can naturally be interpreted as conditional steering. Specifically, the learned low-rank updates can be viewed as input-dependent transformations that steer the model’s hidden representations toward more faithful generation.
\section{Method}

We propose \textbf{T}oken-Level \textbf{O}ff-\textbf{P}olicy \textbf{L}abeling (TOPL, Fig.~\ref{fig:TOPL-model}), a training paradigm that treats post-training as a token-level correctness prediction task. In contrast to next-token prediction, which typically relies on one-hot labels identifying a single correct token, TOPL adopts a binary classification objective that predicts whether each response token is correct or corrupted, where corrupted tokens are tokens that have been perturbed to introduce factual inconsistencies (e.g., in Figure~\ref{fig:TOPL-model}, replacing the correct token “{\color{teal}November}” with “{\color{red}October}” yields a corrupted token). For clarity, we instantiate TOPL using document summarization throughout this section. The same formulation naturally extends to other faithful generation tasks, such as machine translation, which we describe in Section~\ref{MT}.
% When coupled with efficient adaptation methods such as LoRA, TOPL further enables lightweight training while offering improved interpretability. 

 % \gccomment{saying "given the context" is a bit abstact, would be good to just define what a corrupted token is. Something like "...token is correct or corrupted, where a corrupted token is a purposefully unfactual token. For example [insert small example]}

\subsection{Generating Data for Token-Level Supervision}
\label{Generating Token-Level Supervision}
Given a document $D$ and its ground-truth summary $S = (t_1, \ldots, t_{|S|})$, where $t_k$ denotes the $k$-th token and $|S|$ is the total number of tokens in the summary, the sequence S serves as a positive example of a faithful response.

To construct the token-level supervision signal for our method, one natural approach is to generate a perturbed version of the summary, denoted as $S'$, by introducing controlled modifications such as token insertions, deletions, or substitutions that make the content of the summary unfaithful to $D$. These perturbations can affect individual tokens, key phrases, or even entire sentences, enabling precise localization of errors at the token level. In general, such perturbations can be produced by prompting a large language model to generate corrupted summaries from positive examples. In our experiments, we leverage the FAVA dataset \citep{mishra2024fine}, which provides model-generated summaries together with fine-grained edits, i.e. corruptions, to said summary that simulate typical hallucinations. For each token $t_k$ in the perturbed sequence $S'$, we assign a binary label $z_k \in \{0,1\}$, where $z_k = 1$ indicates that the token is faithful to the original summary $S$ (i.e. it is a "good" token), and $z_k = 0$ indicates that the token corresponds to a hallucinated span introduced by the edits (i.e. "bad" tokens).

% \subsection{Training Token-Level Off-Policy Labeling Model}

\subsection{Token-Level Off-Policy Labeling Model}

\paragraph{Training.} Training a TOPL model is conceptually similar to training a reward model, with the key distinction that supervision is provided at the token level rather than at the sequence level. TOPL predicts a correctness score for each response token. Given a document $D$ and a response sequence $S' = (t_1, \ldots, t_{|S'|})$, the model outputs a probability $p_k = P(z_k = 1 \mid D, t_{\le k})$ for every token $t_k$, where $z_k \in \{0,1\}$ denotes the token-level correctness label defined in Section \ref{Generating Token-Level Supervision}.

To produce these predictions, instead of using the decoding head $\ell$, which maps the final hidden states to vocabulary logits for next-token prediction, we attach a lightweight reward head $h : \mathbb{R}^d \rightarrow \mathbb{R}$ at a chosen intermediate layer $m$ of the language model $f$, where $f_{0:m}$ denotes the model truncated up to layer $m$. We first compute the hidden states as:
\begin{align}
H = \mathrm{FinalRMSNorm}\big(f_{0:m}(D, S)\big) \in \mathbb{R}^{|S| \times D_{\text{hidden}}},
\end{align}
where $\mathrm{FinalRMSNorm}(\cdot)$ denotes the final RMS normalization layer of the language model $f$, applied here before the reward head $h$ to ensure consistent scaling with the original decoding head. We then apply $h$ to each token representation:
\begin{align}
P(z_k = 1 \mid D, t_{\le k}) = \sigma\big( h(H_k) \big), \quad k = 1, \dots, |S|,
\end{align}
where $\sigma(\cdot)$ denotes the sigmoid function. The resulting probability represents the model’s estimate of whether token $t_k$ is factually consistent with the input context $D$. The model is trained using a binary cross-entropy loss over all response tokens. Instead of updating the full model parameters, we use Low-Rank Adaptation (LoRA) on the selected layers, and analyze the effect of layer selection in Section~\ref{Layer Selection}. 

\paragraph{Low-Rank Model Merging.} The TOPL objective itself does not explicitly optimize next-token prediction, and the reward head is not used during generation. To transfer these improvements into a generative model after training, we discard the reward head $h$ and the normalization layer introduced before it, and merge the learned LoRA update into $f_{0:m}$ to obtain an updated model $f_{\text{merge}}$. Generation is then performed using the original decoding head $\ell$ with standard forward propagation.

\section{Experiments}

\subsection{Experimental Setup} We primarily evaluate TOPL on document summarization throughout this section, and separately study its cross-task generalization to machine translation in Section~\ref{MT}.

\label{Experimental Setup}
\paragraph{Model Architecture.}
We build our TOPL models on top of Qwen3-8B \citep{yang2025qwen3}, Llama-3.1-8B \citep{grattafiori2024llama}, and Gemma-3-4B \citep{gemma_2025}, which serve as backbone models $f$. We apply LoRA to selected transformer blocks, inserting it across model-specific layer ranges: layers 0--29 for Qwen3-8B, and layers 0--27 for both Llama-3.1-8B and Gemma-3-4B. In all experiments, we use a LoRA rank of $r = 4$ and a scaling hyperparameter $\alpha = 8$. More detailed hyperparameter settings are provided in Appendix~\ref{Model Training Setup and Hyperparameters}.

\paragraph{Evaluation.}
We train TOPL on the FAVA~\citep{mishra2024fine} dataset by splitting the original data into train, validation, and test subsets. The model is trained on the training split and evaluated on the held-out test split for in-distribution (ID) evaluation.  Our primary evaluation focuses on out-of-distribution (OOD) generalization, where we evaluate the model on 11 datasets from AggreFact~\citep{tang2023understanding} using up to 300 unique documents per dataset. Each model (including TOPL and all baselines) generates a summary. We then assess factuality using the Bespoke-MiniCheck-7B evaluation model \citep{tang2024minicheck}, which scores the generated summary conditioned on the input document. The resulting score lies between $[0,1]$, where higher values indicate better factual consistency. For OOD evaluation, we report the average score across all 11 datasets. We consider two evaluation protocols. (1) \textbf{Sentence-level evaluation}: we decompose each generated summary into individual sentences, score each sentence independently, and compute the final score as the average across all sentences. This protocol is used for all main results in the paper. (2) \textbf{Full-summary evaluation}: we directly score the entire summary as a single sequence. Detailed results under this protocol are provided in Appendix~\ref{Full-Summary Evaluation}.

\paragraph{Baselines.} We compare TOPL against Supervised Fine-Tuning (SFT) and Direct Preference Optimization (DPO; ~\citealp{rafailov2023direct}). We further compare against several methods that operate on token-level supervision signals, including Token-Level DPO (TDPO; ~\citealp{zeng2024token}), Token-Level Detective Reward (TLDR; ~\citealp{fu2025tldr}), and Token-Level Unlikelihood Training~\citep{welleck2019neural}. Among these, TLDR is a closely related token-level supervision method originally proposed for vision-language models that applies LoRA across all transformer layers. Finally, we introduce a sequence-level version of our method, which we call Sequence-Level Off-Policy Labeling (SOPL). SOPL mirrors TOPL but operates at the sequence level by optimizing a binary classification objective over the entire summary to predict whether it is factual or not. To create the labels for SOPL, we label each summary with corruptions as 0 and their unaltered counterparts as 1. For a fair comparison, all methods are trained using LoRA with rank $r = 4$. Detailed hyperparameter settings are provided in Appendix~\ref{Model Training Setup and Hyperparameters}.

\subsection{Main Results}

\begin{figure}[t]
\centering
\includegraphics[width=0.75\linewidth]{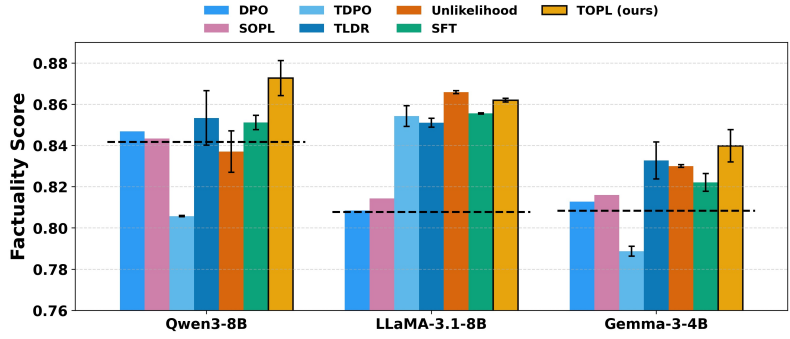}
\caption{
OOD factuality scores across different backbone models. TOPL consistently ranks among the strongest methods and achieves the best average performance across backbone models. Dashed lines indicate the base model performance for each backbone.
}
\label{fig:ood_bar}
\end{figure}

% \begin{figure}[t]
% \centering

% \begin{subfigure}[t]{0.70\linewidth}
%     \centering
%     \includegraphics[width=\linewidth]{ID_plot.pdf}
%     \caption{In-distribution (ID)}
% \end{subfigure}
% \hfill
% \begin{subfigure}[t]{0.70\linewidth}
%     \centering
%     \includegraphics[width=\linewidth]{OOD_plot.pdf}
%     \caption{Out-of-distribution (OOD)}
% \end{subfigure}
% \caption{
% Factuality scores across different backbone models under in-distribution (ID) and out-of-distribution (OOD) settings. 
% TOPL performs best in most settings, and is consistently the top-performing method across all backbones under OOD evaluation. Dashed lines indicate the base model performance for each backbone.
% }
% \label{fig:id_ood_bar}
% \end{figure}

We compare TOPL against SFT, DPO, SOPL, TDPO, TLDR, and Unlikelihood training across both in-distribution (FAVA) and out-of-distribution (AggreFact) benchmarks. 
% The results are shown in Figure~\ref{fig:id_ood_bar}. 
The results are shown in Figure~\ref{fig:ood_bar}. While our primary focus is OOD generalization, we also report in-distribution results on FAVA in Appendix~\ref{numerical results}.

% On in-distribution data, TOPL achieves strong performance across all models, although it does not always outperform all baselines (e.g., on Gemma). Compared to the base model, TOPL achieves substantial improvements, with an average gain of +8.9\%. 
Under distribution shift, TOPL consistently ranks among the strongest methods and achieves the best average performance across backbone models. It obtains the highest scores on Qwen and Gemma and remains competitive with the strongest baseline on Llama. We additionally analyze summary lengths and find that these gains cannot be explained solely by generating shorter or more conservative summaries in Appendix~\ref{Summary Length Analysis}.
% , with an average improvement of +3.6\% over the base model and +1.3\% over SFT (the strongest baseline).

One possible explanation is that, unlike next-token prediction used in SFT, which promotes a single reference token while penalizing all others, TOPL provides a less noisy supervision signal by relaxing the strict one-hot objective. Compared to DPO and SOPL, which rely on sequence-level rewards, TOPL offers finer-grained supervision, contributing to improved performance across out-of-distribution settings. Furthermore, TOPL achieves stronger overall performance than other existing token-level baselines, suggesting that the gains are not solely attributable to token-level supervision itself, but also depend on the particular objective used to incorporate factuality information. Detailed numerical results for each experiment are provided in the Appendix~\ref{numerical results}.
% This demonstrates that the improvements extend beyond in-distribution settings to stronger generalization under distribution shift. 
% \gccomment{I removed the part where you talk about why our method is better, referencing that we don't have a strict one-hot token objective. If we want to add this comment is should be in the discussion. But I am not sure we should since we don't have an experiment to back this hypothesis up. If included, we should say "one possibility for why TOPL outperforms SFT is..."}

\subsection{Layer Selection}
\label{Layer Selection}
\begin{figure}[t]
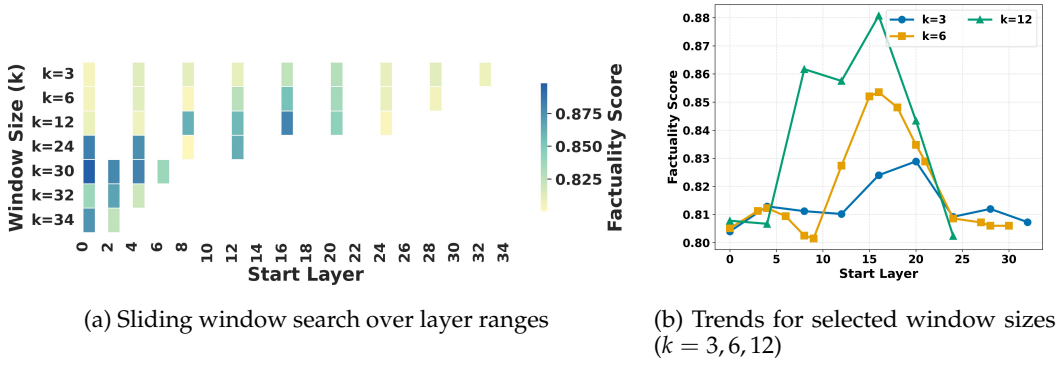

\centering
\begin{subfigure}[t]{0.60\linewidth}
    \centering
    \includegraphics[width=\linewidth]{sliding_window_plt.pdf}
    \caption{Sliding window search over layer ranges}
\end{subfigure}
\hfill
\begin{subfigure}[t]{0.38\linewidth}
    \centering
    \includegraphics[width=\linewidth]{start_layer_small_k.pdf}
    \caption{Trends for selected window sizes ($k=3,6,12$)}
\end{subfigure}

\caption{
Factuality scores under different LoRA insertion ranges.
\textbf{Left:} Sliding window over layer ranges, where darker colors indicate higher scores. 
\textbf{Right:} Plots for representative window sizes ($k=3,6,12$), illustrating that performance peaks in middle layers.
}
\label{fig:sliding_window_combined}
\end{figure}

Since TOPL can be trained with any set of layers of the model, it is not obvious what layers LoRA should be applied. We study this by performing a sliding window search over transformer layers. As shown in Figure~\ref{fig:sliding_window_combined}, applying LoRA to the middle layers yields the most significant factuality improvements, and combining them with early layers further enhances performance. Surprisingly, including the final layers provides limited gains or even worse performance. One possible explanation is that the final layers are more tightly coupled with decoding the hidden state into the output vocabulary; as a result, updating them with an off-policy classification-based objective such as TOPL may disrupt the generative behavior learned during pretraining.
% We further validate this observation in Section~\ref{Interpretability Analysis}.

% % \subsection{Sample Efficiency}
% % \label{Sample Efficiency}

\subsection{Label Sensitivity Analysis}
\label{Label Sensitivity Analysis}
To examine how TOPL responds to variations in its training signals, we conduct a comprehensive label sensitivity analysis, focusing on Qwen3-8B as a case study. By selectively controlling the training tokens and labels, we construct a range of settings that vary both the distribution of labels and the total number of tokens. Specifically, we consider the following configurations: \textit{Original}, which uses the original dataset; \textit{Random}, where token labels are randomly assigned; \textit{Balanced}, where factual (good) and non-factual (bad) tokens are balanced; \textit{Bad-skewed} and \textit{Good-skewed}, where the training distribution is biased toward bad or good tokens, respectively; \textit{Good-only}, where only factual tokens are used for training; and \textit{Dense Balanced}, which maintains a balanced label distribution while increasing the total number of training tokens compared with \textit{Balanced}. Detailed statistics of these datasets, including the number of tokens and label distributions, are provided in Appendix~\ref{Label Sensitivity Dataset Statistics}.

\begin{figure}[t]
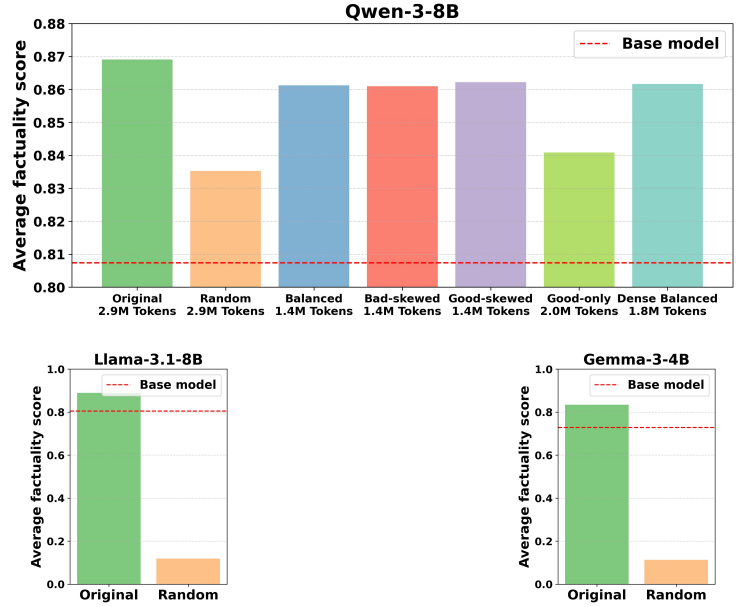

\centering
\begin{subfigure}[t]{0.7\linewidth}
    \centering
    \includegraphics[width=\linewidth]{Qwen_label.pdf}
\end{subfigure}

\vspace{0.5em}
\begin{subfigure}[t]{0.2\linewidth}
    \centering
    \includegraphics[width=\linewidth]{Llama_label.pdf}
\end{subfigure}
\hspace{0.25\linewidth}
\begin{subfigure}[t]{0.2\linewidth}
    \centering
    \includegraphics[width=\linewidth]{Gemma_label.pdf}
\end{subfigure}
\caption{
Factuality scores under different label settings. \textbf{Top:} Qwen3-8B results under various data configurations. With a similar number of training tokens, varying the ratio of good and bad tokens has only a minor effect, while using only one type leads to noticeable degradation. \textbf{Bottom left:} \textit{Original} vs. \textit{Random} settings on Llama-3.1-8B. \textbf{Bottom right:} \textit{Original} vs. \textit{Random} settings on Gemma-3-4B. Random labeling substantially degrades performance for both models, in contrast to Qwen3-8B (top), which still outperforms the base model.
}
\label{fig:label_sensitivity}
\end{figure}

As shown in Figure~\ref{fig:label_sensitivity} (top), the \textit{Balanced}, \textit{Bad-skewed}, and \textit{Good-skewed} settings use a comparable number of training tokens but differ in label distributions. We find that skewing the training set toward either good or bad tokens leads to only minor changes in performance, suggesting that the label ratio is not the dominant factor. In contrast, although \textit{Good-only} and \textit{Dense Balanced} involve a similar number of tokens (with \textit{Good-only} being slightly larger), \textit{Dense Balanced} performs significantly better. This indicates that, while precise balancing is not critical, the presence of both positive and negative supervision is essential. In other words, contrastive supervision plays a key role in shaping model behavior.

We next examine the effect of random labeling. As shown in Figure~\ref{fig:label_sensitivity} (bottom), random labeling substantially degrades performance for Llama-3.1-8B and Gemma-3-4B. However, as shown in Figure~\ref{fig:label_sensitivity} (top), Qwen3-8B surprisingly achieves improvements over the base model even under random labels. This is consistent with prior findings of Qwen3-8B's robustness to spurious supervision~\citep{shao2025spurious}.

\subsection{Generalization to Machine Translation}
\label{MT}

\begin{wrapfigure}{r}{0.4\textwidth}
\centering
\vspace{-7ex}
\includegraphics[width=\linewidth]{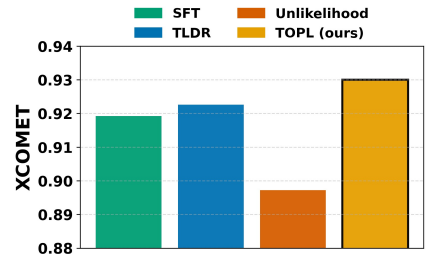}
\caption{
Machine translation results on Qwen3-8B measured by XCOMET (higher is better). 
TOPL achieves the best performance among all methods on the out-of-distribution (OOD) FLORES+ benchmark.
}
\label{fig:translation_results}
\vspace{-4ex}
\end{wrapfigure}

To evaluate whether TOPL extends to other faithful generation tasks, we additionally apply it to machine translation. We train Qwen3-8B on four language pairs from MLQE-PE~\citep{fomicheva2022mlqe} (en-de, en-zh, ro-en, and et-en), where word-level quality annotations are converted into token-level supervision signals. We evaluate on both the held-out MLQE-PE test set (ID) and FLORES-Plus~\citep{nllb-24, maillard-etal-2024-findings} (OOD) using the same language pairs and report XCOMET scores~\citep{guerreiro2024xcomet}, where higher values indicate better translation quality.

As shown in Figure~\ref{fig:translation_results}, TOPL achieves the strongest performance among all compared methods on out-of-distribution benchmarks. Notably, the gains persist despite substantial differences from summarization, including the supervision source, evaluation metric, and output structure. These results suggest that the benefits of TOPL are not specific to summarization and may extend to other faithful generation tasks where token-level supervision is available. Full results, including in-distribution evaluation, are provided in Appendix~\ref{appendix_translation_results}.

\section{Analysis}
\label{Analysis}

\subsection{TOPL with LoRA as Conditional Steering}
\label{TOPL with LoRA as Conditional Steering}

TOPL with LoRA can be viewed as a form of conditional steering~\citep{lee2025programming}, where a hidden state $h \in \mathbb{R}^d$ is modified along a steering direction $v$ with strength determined by its alignment with a concept vector $c$:
\begin{align}
    h' &\leftarrow h + \alpha \cdot g(\mathrm{sim}(h, \mathrm{proj}_c h)) \cdot v.
\end{align}

Suppose the gate function $g$ is identity, and both $c$ and $h$ are unit-normalized. Then the update can be written as a \textit{rank-1} transformation:
\begin{equation}
    h' = h + \alpha \cdot h^\top \left(cc^\top h\right) \cdot v
       = h + \tau \alpha \cdot (v c^\top)\, h,
\end{equation}
where $\tau = h^\top c$ and $v c^\top$ defines the \textit{rank-1} update. Similarly, a LoRA-adapted layer can be written as
\begin{align}
h' = W h + \alpha \sum_{i=1}^r v_i (c_i^\top h),
\end{align}
where $\{c_i\}_{i=1}^r$ and $\{v_i\}_{i=1}^r$ correspond to the rows of $A$ and columns of $B$. Each term $v_i (c_i^\top h)$ matches the conditional steering form, with $c_i^\top h$ acting as the input-dependent steering strength and $v_i$ as the steering direction. Consequently, the LoRA $A$ matrix encodes projections onto concept directions (conditioning), while the LoRA $B$ matrix defines the corresponding steering directions.

Based on this correspondence, we hypothesize that TOPL with LoRA can be interpreted as learning multiple adaptive steering mechanisms. Given an input hidden state, the LoRA-$A$ acts as the conditioning component that projects the representation onto a set of learned concept directions, which can be interpreted as detecting whether the current token is aligned with factual or non-factual patterns. The resulting responses determine how each corresponding LoRA-$B$ direction is activated, where LoRA-$B$ can be interpreted as a set of learned steering directions that modulate the hidden representation toward or away from factual behavior. We empirically validate this interpretation through ablation studies of the LoRA components for TOPL and SFT on document summarization, analyzing the distinct roles of $A$ and $B$ in Section~\ref{LoRA-$A$ as a Token-Level Classifier} and Section~\ref{LoRA-$B$ as Steering Vectors}. 
% We further provide corresponding analyses for DPO and SOPL in Appendix~\ref{appendix_DPO&SOPL}.

% Unlike standard steering methods that rely on fixed or manually designed directions, TOPL jointly learns these concept-direction pairs across layers in a data-driven manner, enabling more flexible and distributed control of model behavior.

\subsection{LoRA-$A$ as a Token-Level Classifier}
\label{LoRA-$A$ as a Token-Level Classifier}

We hypothesize that the LoRA-$A$ component captures discriminative concept directions for token correctness, effectively separating factual and nonfactual tokens in the hidden representation space. To validate this, we project hidden states onto the LoRA-$A$ subspace and examine whether the good and bad tokens can be separated. This analysis is performed across all layers with LoRA modules. We quantify separability using the Area Under the ROC curve (AUROC), which measures how well the projected representations distinguish between good and bad tokens. This analysis is performed for both TOPL and SFT to compare how well LoRA-$A$ captures token-level correctness under different training objectives.

As shown in Figure~\ref{fig:loraA_loraB_compare} (left), TOPL exhibits substantially stronger separability between good and bad tokens in the LoRA-$A$ subspace compared to SFT. This suggests that TOPL and SFT leverage LoRA in fundamentally different ways. The separability becomes increasingly pronounced in the middle-to-late layers, while earlier layers show weaker separation. One possible explanation is that early layers are still processing lower level information from the input representations, whereas middle-to-late layers are more involved in extracting and manipulating higher level features, such as factuality, for downstream generation, making them more directly engaged in conditional steering.

\subsection{LoRA-$B$ as Steering Vectors}
\label{LoRA-$B$ as Steering Vectors}

We next study the role of LoRA-$B$, which we interpret as encoding steering directions associated with factuality. To this end, the learned LoRA updates are merged into the backbone model, followed by post-hoc interventions along LoRA-$B$ directions, restricted to modules identified by the LoRA-$A$ analysis as exhibiting strong separability. Specifically, we compute the difference between the mean representations of good and bad tokens, denoted by $\mu_{+}-\mu_{-}$, and map this difference through LoRA-$B$ to obtain a steering update. We then apply interventions of the form
\begin{align}
\Delta h = \lambda (\mu_{+}-\mu_{-}) B,
\end{align}
The sign of $\lambda$ controls the direction of the intervention. Accordingly, $(\mu_{+}-\mu_{-})B$ can be interpreted as a direction that promotes factual behavior, with its negative corresponding to movement toward hallucinated behavior. 

% \gccomment{I think using $\alpha$ here can be a bit confusing since it is a hyperparameter earlier on... this could also apply to the analysis section. Maybe use another symbol?}
% In particular, when a token is aligned with factual patterns, LoRA-$B$ steers the representation further toward directions associated with factual behavior, while the opposite direction corresponds to hallucinated content.

\begin{figure}[t]
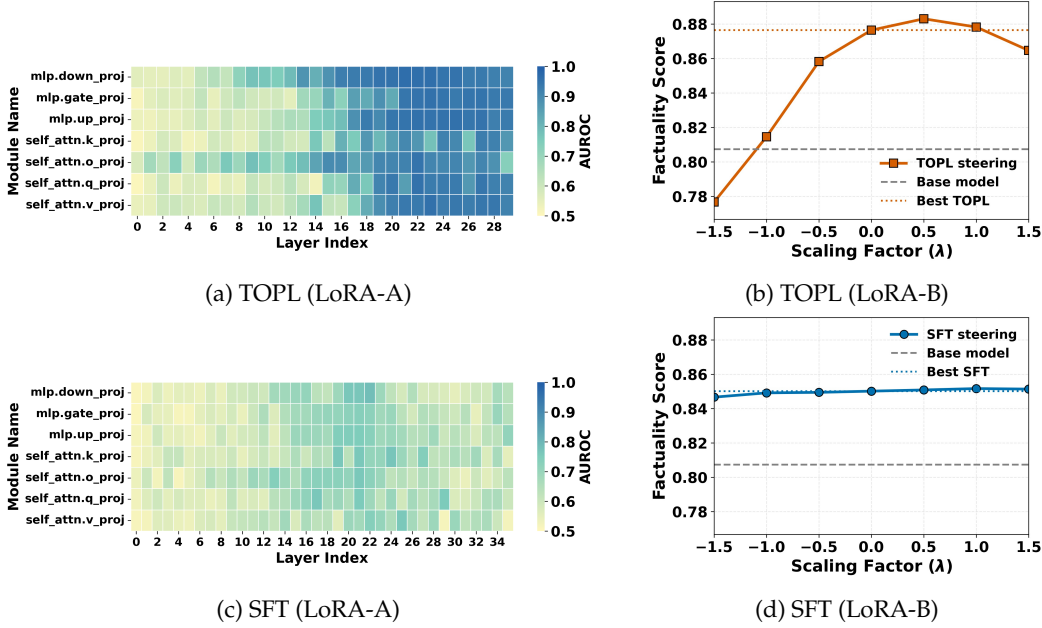

\centering
% ===== Row 1: TOPL =====
\begin{subfigure}[t]{0.60\linewidth}
    \centering
    \includegraphics[width=\linewidth]{loraA_TOPL.pdf}
    \caption{TOPL (LoRA-A)}
\end{subfigure}
\hfill
\begin{subfigure}[t]{0.38\linewidth}
    \centering
    \includegraphics[width=\linewidth]{loraB_TOPL.pdf}
    \caption{TOPL (LoRA-B)}
\end{subfigure}
\vspace{0.5em}
% ===== Row 2: SFT =====
\begin{subfigure}[t]{0.60\linewidth}
    \centering
    \includegraphics[width=\linewidth]{loraA_SFT.pdf}
    \caption{SFT (LoRA-A)}
\end{subfigure}
\hfill
\begin{subfigure}[t]{0.38\linewidth}
    \centering
    \includegraphics[width=\linewidth]{loraB_SFT.pdf}
    \caption{SFT (LoRA-B)}
\end{subfigure}
\caption{Comparison of LoRA-based representations across training methods and parameter types.
\textbf{Top row:} TOPL; \textbf{bottom row:} SFT.
\textbf{Left column:} LoRA-A; \textbf{right column:} LoRA-B.
LoRA-A heatmaps (left) show the separability (AUROC) between factual and non-factual tokens across layers and modules, where higher values indicate stronger discriminative structure. TOPL exhibits substantially higher separability in the middle and later layers compared to SFT.
LoRA-B plots (right) show the effect of scaling the LoRA updates, illustrating how different training methods respond to steering strength. The lines labeled as “best TOPL” and “best SFT” correspond to the models obtained after merging all LoRA updates (i.e., before applying any steering). TOPL displays a clear and consistent response to the scaling factor $\lambda$.
}
\label{fig:loraA_loraB_compare}
\end{figure}

As shown in Figure~\ref{fig:loraA_loraB_compare} (right), for TOPL, factuality first increases and then decreases as $\lambda$ varies from negative to positive values. This suggests that negative $\lambda$ steers the model toward nonfactual directions, while positive $\lambda$ steers it toward factual representations, with an optimal magnitude balancing these effects. In contrast, SFT exhibits much weaker sensitivity to such interventions, suggesting that its updates do not encode a strong coherent steering direction.
% In contrast, SFT exhibits much weaker sensitivity to such interventions, suggesting that its updates do not encode a similarly coherent steering direction, and are therefore less interpretable. 
DPO and SOPL similarly exhibit relatively weak LoRA-$A$ separability but can exhibit clear LoRA-$B$ steering effects; the corresponding analyses are provided in Appendix~\ref{appendix_DPO&SOPL}.

\subsection{LoRA-$A$ as an Indicator of OOD Generalization}
\label{LoRA-A as a Predictor of OOD Performance}
% We investigate the relationship between representation quality and generalization under distribution shift. As summarized in Table~\ref{tab:loraA_ood_qwen}, we observe an alignment: methods that achieve stronger separability in the LoRA-$A$ subspace tend to exhibit better OOD performance. In particular, TOPL attains substantially higher AUROC in LoRA-$A$, while also achieving the strongest factuality scores on OOD datasets. This alignment suggests a possible explanation: LoRA-$A$ may capture directions that distinguish between factual and hallucinated tokens, effectively acting as a low-rank projection that conditions hidden states on token-level correctness. When this separation is strong, the model is more likely to rely on stable semantic features rather than superficial patterns tied to the training distribution. In contrast, when factual and hallucinated tokens are less separable in the representation space (i.e., lower AUROC), the model is more prone to rely on spurious correlations, which may not generalize under distribution shift. This leads to degraded performance on OOD inputs. While we do not claim a strict causal relationship, our results indicate that LoRA-$A$ separability serves as a useful representation-level indicator for understanding and diagnosing generalization behavior in post-trained language models.

We investigate the relationship between LoRA-$A$'s separability and generalization under distribution shift. As shown in Table~\ref{tab:loraA_ood_qwen}, methods with stronger separability in the LoRA-$A$ subspace tend to achieve better OOD performance. In particular, TOPL exhibits both higher AUROC in LoRA-$A$ and stronger factuality on OOD datasets. This might be because of LoRA-$A$ captures directions distinguishing factual from non-factual tokens, acting as a low-rank projection that conditions hidden states on token-level correctness. When this separation is weak, the model is more likely to rely on spurious correlations, leading to degraded OOD performance. While we do not claim a strict causal relationship, our results indicate that LoRA-$A$ separability serves as a useful representation-level indicator for understanding generalization behavior in post-trained language models.

\begin{table}[t]
\centering
\begin{tabular}{lccc}
\toprule
Method & LoRA-$A$ AUROC & OOD Score \\
\midrule
SOPL & 0.6064 $\pm$ 0.0648 & 0.8434 \\
DPO & 0.6264 $\pm$ 0.0682 & 0.8473 \\
SFT & 0.6272 $\pm$ 0.0666 & 0.8530 \\
\textbf{TOPL (ours)} & \textbf{0.7541 $\pm$ 0.1573} & \textbf{0.8706} \\
\bottomrule
\end{tabular}
\caption{
Comparison of LoRA-$A$ separability and out-of-distribution performance on Qwen3-8B. 
TOPL achieves significantly higher separability in the LoRA-$A$ subspace, which aligns with better generalization under distribution shift.
}
\label{tab:loraA_ood_qwen}
\end{table}

\section{Discussion and Conclusion}

% In this work, we introduced Token-Level Supervision (TOPL), a novel off-policy method that reframes language model post-training as a token-level, correctness prediction task. We show that TOPL consistently improves factuality on summarization tasks under both in-distribution and out-of-distribution settings compared to sequence-level supervision and supervised fine-tuning. We further provide a mechanistic interpretation of TOPL with LoRA, showing that LoRA-$A$ and LoRA-$B$ decompose into a conditional steering framework, corresponding to concept projection and steering directions. Our analysis reveals a consistent relationship between LoRA-$A$ separability and OOD performance, suggesting that representation quality may serve as an indicator of generalization. We also study the sensitivity of TOPL to different training set distributions and find that its effectiveness depends on the underlying model. Overall, TOPL offers a simple, effective, and interpretable alternative to existing post-training methods for faithful summary generation.

In this work, we introduced Token-Level Off-Policy Labeling (TOPL), an off-policy post-training framework that reframes language model alignment as token-level correctness prediction. We show that TOPL consistently improves factuality on document summarization, generalizes to machine translation, and achieves strong out-of-distribution robustness compared to both sequence-level and token-level supervision methods.

There are several possible hypotheses for why TOPL may generalize more effectively than conventional approaches. Compared to sequence-level supervision, TOPL provides finer-grained training signals by localizing supervision to individual tokens, resulting in substantially richer supervision information. However, our comparisons with other token-level methods suggest that the gains cannot be attributed solely to the presence of token-level supervision. A key distinction of TOPL is that it does not optimize a generation objective directly. Instead, the model is trained to distinguish correct and incorrect tokens conditioned on context, indirectly reshaping the intermediate representation space. Our mechanistic analysis suggests that this process encourages more separable internal representations between faithful and hallucinated behaviors, particularly within the LoRA adaptation subspace. Rather than memorizing surface token distributions, TOPL appears to promote reliance on more stable and semantically grounded features, which may explain its stronger robustness under distribution shift.

While TOPL shows strong empirical performance, several directions remain for future work. First, improving data quality by generating cleaner token-level perturbations could potentially reduce supervision noise and further enhance performance. Second, TOPL may extend naturally to reasoning tasks, where correctness depends on intermediate steps and TOPL can capture fine-grained errors. Third, varying token-level labels could enable more flexible control over model behavior beyond factuality. Finally, although we observe a correlation between LoRA-$A$ separability and OOD performance, understanding the causal relationship between representation structure and generalization remains an open question.

% \section*{Author Contributions}

% \section*{Ethics Statement}
% Authors can add an optional ethics statement to the paper. 
% For papers that touch on ethical issues, this section will be evaluated as part of the review process. The ethics statement should come at the end of the paper. It does not count toward the page limit, but should not be more than 1 page. 

\section*{Acknowledgments}
We thank Wang Bill Zhu and Ameya Godbole for the insightful discussions and valuable feedback that contributed to this work.
DF was supported in part by a gift from the USC - Capital One Center for Responsible AI and Decision Making in Finance (CREDIF).
This work was supported in part by the National Science Foundation under Grant No. IIS-2403436. Any opinions, findings, and conclusions or recommendations expressed in this material are those of the author(s) and do not necessarily reflect the views of the National Science Foundation.
This research is based upon work supported in part by the Office of the Director of National Intelligence (ODNI), Intelligence Advanced Research Projects Activity (IARPA), via 56000026C0020. The views and conclusions contained herein are those of the authors and should not be interpreted as necessarily representing the official policies, either expressed or implied, of ODNI, IARPA, or the U.S. Government. The U.S. Government is authorized to reproduce and distribute reprints for governmental purposes notwithstanding any copyright annotation therein.

\bibliography{main}
\bibliographystyle{colm2026_conference}

\clearpage
\appendix
\section{Appendix}

% \subsection{Normalization Layer Placement}
% \label{Normalization Layer Placement}
% In practice, when training TOPL, we modify the placement of the final normalization layer by moving it to the output 
% of the truncated model $f_{0:m}$, i.e., immediately before the reward head $h$.

% Formally, let $\mathrm{Norm}(\cdot)$ denote the final normalization layer of the pretrained  model $f$. We compute the hidden states as
% \begin{align}
% H = \mathrm{Norm}\big(f_{0:m}(D, S)\big),
% \end{align}
% and apply the reward head $h$ on the normalized representations.

% This modification ensures that the input to the reward head is consistent with the scale 
% and distribution of hidden states used by the original decoding head $\ell$, leading to 
% more stable training.

\subsection{Full-Summary Evaluation}
\label{Full-Summary Evaluation}

In addition to the sentence-level evaluation used in the main paper, we consider an alternative evaluation protocol that scores the entire generated summary as a single sequence. Specifically, instead of decomposing the summary into individual sentences, we directly evaluate the full summary using the Bespoke-MiniCheck-7B model.

\begin{figure}[H]
    \centering
    \includegraphics[width=0.6\linewidth]{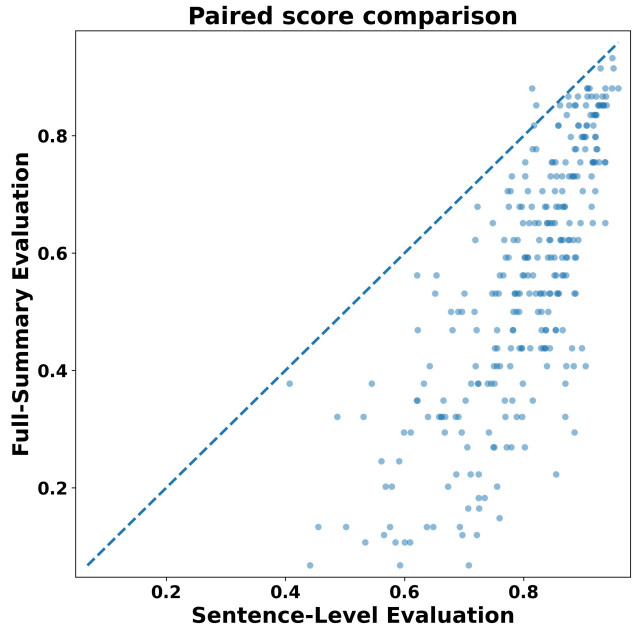}
    \caption{Comparison between sentence-level and full-summary evaluation scores on 300 samples. Each point represents a generated summary. The dashed line indicates equality. Sentence-level evaluation generally assigns higher scores, while maintaining a strong rank correlation (Spearman = 0.79).}
    \label{fig:full_vs_sentence}
\end{figure}

We first examine the relationship between full-summary evaluation and the sentence-level evaluation. To this end, we evaluate 300 samples using both protocols and compare their scores. As shown in Figure~\ref{fig:full_vs_sentence}, sentence-level evaluation tends to assign slightly higher scores in most cases. This is likely because decomposing summaries into individual sentences allows well-formed or factual sentences to raise the overall score, even if other parts of the summary are less accurate. Despite this difference in absolute values, the two evaluation protocols exhibit a strong rank correlation, with a Spearman correlation coefficient of 0.79 across the 300 samples. This suggests that the relative ordering of model outputs is largely preserved.

% We further evaluate some methods as an example under the full-summary evaluation protocol on both in-distribution (ID) and out-of-distribution (OOD) datasets. As shown in Figure~\ref{fig:full_summary_id_ood}, under in-distribution evaluation, TOPL achieves the best performance on Qwen3-8B and Llama-3.1-8B, while on Gemma-3-4B it is competitive but not the top-performing method. In contrast, under out-of-distribution evaluation, TOPL consistently outperforms all baselines across all backbone models, demonstrating strong generalization capability. These results are consistent with those observed under sentence-level evaluation, further validating the robustness of our conclusions across different evaluation protocols.

\begin{figure}[H]
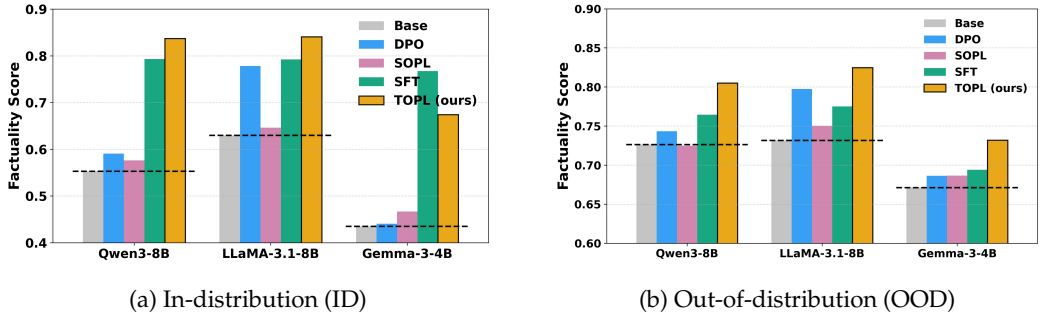

\centering
\begin{subfigure}[t]{0.48\linewidth}
    \centering
    \includegraphics[width=\linewidth]{ID_plot_fullsum.pdf}
    \caption{In-distribution (ID)}
\end{subfigure}
\hfill
\begin{subfigure}[t]{0.48\linewidth}
    \centering
    \includegraphics[width=\linewidth]{OOD_plot_fullsum.pdf}
    \caption{Out-of-distribution (OOD)}
\end{subfigure}
\caption{
Factuality scores under full-summary evaluation across different backbone models in both in-distribution (ID) and out-of-distribution (OOD) settings. Dashed lines indicate the base model performance for each backbone.
}
\label{fig:full_summary_id_ood}
\end{figure}

We further evaluate a representative subset of methods under the full-summary evaluation protocol on both in-distribution (ID) and out-of-distribution (OOD) datasets. As shown in Figure~\ref{fig:full_summary_id_ood}, under in-distribution evaluation, TOPL achieves the best performance on Qwen3-8B and LLaMA-3.1-8B. Under out-of-distribution evaluation, TOPL consistently achieves the strongest performance among the methods considered across all backbone models. Overall, the trends closely mirror those observed under sentence-level evaluation, suggesting that our conclusions are robust to the choice of evaluation protocol.

\subsection{Summary Length Analysis}
\label{Summary Length Analysis}

A potential concern is that the factuality gains may arise simply from generating shorter or more conservative summaries. To investigate this possibility, we compare the average summary lengths of TOPL and SFT method under OOD evaluation as an example.

As shown in Table~\ref{tab:length_analysis}, TOPL often produces shorter summaries than SFT on Qwen and Llama, but this trend is not universal. On Gemma, for example, TOPL generates slightly longer summaries than SFT while still achieving higher factuality. Moreover, compared to the base model, TOPL generates slightly more sentences on both Qwen and Gemma, yet yields markedly larger improvements in factuality. These results suggest that summary length alone cannot explain the observed improvements.

\begin{table}[H]
\centering
\small
\begin{tabular}{lccc}
\toprule
& Qwen & Llama & Gemma \\
\midrule
\multicolumn{4}{c}{\textbf{OOD Average Factuality Score}} \\
\midrule
TOPL (Ours) & \textbf{0.8726 $\pm$ 0.0085} & \textbf{0.8619 $\pm$ 0.0009} & \textbf{0.8398 $\pm$ 0.0078} \\
SFT & 0.8511 $\pm$ 0.0035 & 0.8555 $\pm$ 0.0003 & 0.8221 $\pm$ 0.0043 \\
Base & 0.8417 & 0.8077 & 0.8083 \\
\midrule
\multicolumn{4}{c}{\textbf{OOD Average Sentence Count}} \\
\midrule
TOPL (Ours) & \textbf{2.9894 $\pm$ 0.5528} & \textbf{2.4773 $\pm$ 0.3548} & \textbf{4.6445 $\pm$ 0.4163} \\
SFT & 4.4867 $\pm$ 0.0839 & 3.7258 $\pm$ 0.1139 & 4.4821 $\pm$ 0.0340 \\
Base & 2.3491 & 3.3845 & 3.4627 \\
\bottomrule
\end{tabular}
\caption{OOD factuality scores and average sentence counts. On Gemma, TOPL generates slightly longer summaries while still achieving higher factuality, suggesting that summary length alone does not explain the observed gains.}
\label{tab:length_analysis}
\end{table}

\subsection{Model Training Setup and Hyperparameters}
\label{Model Training Setup and Hyperparameters}

We provide the model architectures, training setups, and hyperparameter configurations for TOPL, SFT, SOPL, DPO, TLDR, Token-level Unlikelihood Training, and Token-Level DPO in Tables~\ref{tab:topl_hparams}, \ref{tab:sft_hparams}, \ref{tab:sopl_hparams}, \ref{tab:dpo_hparams}, \ref{tab:tldr_hparams}, \ref{tab:unl_hparams}, and \ref{tab:tdpo_hparams}.

\begin{table}[H]
\centering
\small
\begin{tabular}{lccc}
\toprule
\multicolumn{4}{c}{\textbf{TOPL Training Hyperparameters}} \\
\midrule
 & \textbf{Qwen3-8B} & \textbf{Llama-3.1-8B} & \textbf{Gemma-3-4B} \\
\midrule
LoRA Rank ($r$) & 4 & 4 & 4 \\
LoRA $\alpha$ & 8 & 8 & 8 \\
LoRA Dropout & 0.1 & 0.1 & 0.1 \\
GPU & \multicolumn{3}{c}{1 $\times$ RTX A6000} \\
Batch Size (per device) & 16 & 16 & 16 \\
Gradient Accumulation & 1 & 1 & 1 \\
Learning Rate & $5\times10^{-5}$ & $1\times10^{-4}$ & $1\times10^{-4}$ \\
LoRA Start Layer & 0 & 0 & 0 \\
LoRA End Layer & 29 & 27 & 27 \\
\bottomrule
\end{tabular}
\caption{Training hyperparameters for TOPL across different backbone models.}
\label{tab:topl_hparams}
\end{table}

\begin{table}[H]
\centering
\small
\begin{tabular}{lccc}
\toprule
\multicolumn{4}{c}{\textbf{SFT Training Hyperparameters}} \\
\midrule
 & \textbf{Qwen3-8B} & \textbf{Llama-3.1-8B} & \textbf{Gemma-3-4B} \\
\midrule
LoRA Rank ($r$) & 4 & 4 & 4 \\
LoRA $\alpha$ & 8 & 8 & 8 \\
LoRA Dropout & 0.1 & 0.1 & 0.1 \\
GPU & \multicolumn{3}{c}{4 $\times$ RTX A6000} \\
Batch Size (per device) & 4 & 4 & 4 \\
Gradient Accumulation & 4 & 4 & 4 \\
Learning Rate & $1\times10^{-4}$ & $1\times10^{-4}$ & $5\times10^{-5}$ \\
LoRA Start Layer & 0 & 0 & 0 \\
LoRA End Layer & 35 & 31 & 33 \\
\bottomrule
\end{tabular}
\caption{Training hyperparameters for SFT across different backbone models.}
\label{tab:sft_hparams}
\end{table}

\begin{table}[H]
\centering
\small
\begin{tabular}{lccc}
\toprule
\multicolumn{4}{c}{\textbf{SOPL Training Hyperparameters}} \\
\midrule
 & \textbf{Qwen3-8B} & \textbf{Llama-3.1-8B} & \textbf{Gemma-3-4B} \\
\midrule
LoRA Rank ($r$) & 4 & 4 & 4 \\
LoRA $\alpha$ & 8 & 8 & 8 \\
LoRA Dropout & 0.1 & 0.1 & 0.1 \\
GPU & \multicolumn{3}{c}{4 $\times$ RTX A6000} \\
Batch Size (per device) & 8 & 8 & 8 \\
Gradient Accumulation & 1 & 1 & 1 \\
Learning Rate & $1\times10^{-4}$ & $1\times10^{-4}$ & $1\times10^{-4}$ \\
LoRA Start Layer & 0 & 0 & 0 \\
LoRA End Layer & 35 & 27 & 33 \\
\bottomrule
\end{tabular}
\caption{Training hyperparameters for SOPL across different backbone models.}
\label{tab:sopl_hparams}
\end{table}

\begin{table}[H]
\centering
\small
\begin{tabular}{lccc}
\toprule
\multicolumn{4}{c}{\textbf{DPO Training Hyperparameters}} \\
\midrule
 & \textbf{Qwen3-8B} & \textbf{Llama-3.1-8B} & \textbf{Gemma-3-4B} \\
\midrule
LoRA Rank ($r$) & 4 & 4 & 4 \\
LoRA $\alpha$ & 8 & 8 & 8 \\
LoRA Dropout & 0.0 & 0.0 & 0.0 \\
GPU & \multicolumn{3}{c}{4 $\times$ RTX A6000} \\
Batch Size (per device) & 1 & 1 & 1 \\
Gradient Accumulation & 8 & 8 & 8 \\
Learning Rate & $5\times10^{-7}$ & $1\times10^{-6}$ & $5\times10^{-7}$ \\
LoRA Start Layer & 0 & 0 & 0 \\
LoRA End Layer & 35 & 31 & 33 \\
\bottomrule
\end{tabular}
\caption{Training hyperparameters for DPO across different backbone models.}
\label{tab:dpo_hparams}
\end{table}

\begin{table}[H]
\centering
\small
\begin{tabular}{lccc}
\toprule
\multicolumn{4}{c}{\textbf{TLDR Training Hyperparameters}} \\
\midrule
 & \textbf{Qwen3-8B} & \textbf{Llama-3.1-8B} & \textbf{Gemma-3-4B} \\
\midrule
LoRA Rank ($r$) & 4 & 4 & 4 \\
LoRA $\alpha$ & 8 & 8 & 8 \\
LoRA Dropout & 0.1 & 0.1 & 0.1 \\
GPU & \multicolumn{3}{c}{4 $\times$ RTX A6000} \\
Batch Size (per device) & 16 & 16 & 16 \\
Gradient Accumulation & 1 & 1 & 1 \\
Learning Rate & $1\times10^{-4}$ & $1\times10^{-4}$ & $1\times10^{-4}$ \\
LoRA Start Layer & 0 & 0 & 0 \\
LoRA End Layer & 35  & 31 & 33 \\
\bottomrule
\end{tabular}
\caption{Training hyperparameters for TLDR across different backbone models.}
\label{tab:tldr_hparams}
\end{table}

\begin{table}[H]
\centering
\small
\begin{tabular}{lccc}
\toprule
\multicolumn{4}{c}{\textbf{Unlikelihood Training Hyperparameters}} \\
\midrule
 & \textbf{Qwen3-8B} & \textbf{Llama-3.1-8B} & \textbf{Gemma-3-4B} \\
\midrule
LoRA Rank ($r$) & 4 & 4 & 4 \\
LoRA $\alpha$ & 8 & 8 & 8 \\
LoRA Dropout & 0.1 & 0.1 & 0.1 \\
GPU & \multicolumn{3}{c}{4 $\times$ RTX A6000} \\
Batch Size (per device) & 4 & 4 & 4 \\
Gradient Accumulation & 2 & 2 & 2 \\
Learning Rate & $1\times10^{-4}$ & $1\times10^{-4}$ & $1\times10^{-4}$ \\
LoRA Start Layer & 0 & 0 & 0 \\
LoRA End Layer & 35  & 31 & 33 \\
\bottomrule
\end{tabular}
\caption{Training hyperparameters for Token-Level Unlikelihood Training across different backbone models.}
\label{tab:unl_hparams}
\end{table}

\begin{table}[H]
\centering
\small
\begin{tabular}{lccc}
\toprule
\multicolumn{4}{c}{\textbf{TDPO Training Hyperparameters}} \\
\midrule
 & \textbf{Qwen3-8B} & \textbf{Llama-3.1-8B} & \textbf{Gemma-3-4B} \\
\midrule
LoRA Rank ($r$) & 4 & 4 & 4 \\
LoRA $\alpha$ & 8 & 8 & 8 \\
LoRA Dropout & 0.1 & 0.1 & 0.1 \\
GPU & \multicolumn{3}{c}{2 $\times$ RTX A6000} \\
Batch Size (per device) & 4 & 4 & 4 \\
Gradient Accumulation & 4 & 4 & 4 \\
Learning Rate & $5\times10^{-6}$ & $5\times10^{-6}$ & $5\times10^{-6}$ \\
LoRA Start Layer & 0 & 0 & 0 \\
LoRA End Layer & 35  & 31 & 33 \\
\bottomrule
\end{tabular}
\caption{Training hyperparameters for TDPO Training across different backbone models.}
\label{tab:tdpo_hparams}
\end{table}

\subsection{Detailed Results on In-Distribution and Out-of-Distribution Benchmarks}
\label{numerical results}
This section provides detailed numerical results for all models and methods under both in-distribution (ID) and out-of-distribution (OOD) settings. Tables~\ref{tab:id_results} and~\ref{tab:ood_results} report the corresponding results.

\begin{table}[!htbp]
\centering
\small
\begin{tabular}{lccc}
\toprule
Method & Qwen3-8B & LLaMA-3.1-8B & Gemma-3-4B \\
\midrule
TOPL (Ours) & \textbf{0.8687 $\pm$ 0.0198} & \textbf{0.8710 $\pm$ 0.0070} & \textbf{0.8119 $\pm$ 0.0202} \\
SFT & 0.8431 $\pm$ 0.0030 & 0.8436 $\pm$ 0.0030 & 0.8368 $\pm$ 0.0081 \\
TLDR & 0.8364 $\pm$ 0.0119 & 0.8761 $\pm$ 0.0134 & 0.7829 $\pm$ 0.0405 \\
TDPO & 0.8304 $\pm$ 0.0019 & 0.8536 $\pm$ 0.0014 & 0.8299 $\pm$ 0.0024 \\
Unlikelihood & 0.8375 $\pm$ 0.0024 & 0.8419 $\pm$ 0.0017 & 0.8342 $\pm$ 0.0025 \\
SOPL & 0.8116 & 0.7932 & 0.7359 \\
DPO & 0.8493 & 0.8441 & 0.8478 \\
\bottomrule
\end{tabular}
\caption{In-distribution factuality scores across backbone models.}
\label{tab:id_results}
\end{table}

\begin{table}[!htbp]
\centering
\small
\begin{tabular}{lccc}
\toprule
Method & Qwen3-8B & LLaMA-3.1-8B & Gemma-3-4B \\
\midrule
TOPL (Ours) & \textbf{0.8726 $\pm$ 0.0085} & \textbf{0.8619 $\pm$ 0.0009} & \textbf{0.8398 $\pm$ 0.0078} \\
SFT & 0.8511 $\pm$ 0.0035 & 0.8555 $\pm$ 0.0003 & 0.8221 $\pm$ 0.0043 \\
TLDR & 0.8533 $\pm$ 0.0132 & 0.8510 $\pm$ 0.0022 & 0.8327 $\pm$ 0.0090 \\
TDPO & 0.8057 $\pm$ 0.0003 & 0.8542 $\pm$ 0.0050 & 0.7887 $\pm$ 0.0024 \\
Unlikelihood & 0.8370 $\pm$ 0.0101 & 0.8658 $\pm$ 0.0007 & 0.8300 $\pm$ 0.0007 \\
SOPL & 0.8434 & 0.8143 & 0.8160 \\
DPO & 0.8473 & 0.8563 & 0.8123 \\
\bottomrule
\end{tabular}
\caption{Out-of-distribution factuality scores averaged across the 11 AggreFact datasets.}
\label{tab:ood_results}
\end{table}

\subsection{Label Sensitivity Dataset Statistics}
\label{Label Sensitivity Dataset Statistics}

We report the statistics of the datasets used in the label sensitivity analysis, including the number of supervised tokens and the distribution of positive and negative labels for each setting in Table~\ref{tab:label_stats}.

\begin{table}[H]
\centering
\small
\begin{tabular}{lccc}
\toprule
\textbf{Setting} & \textbf{\#Tokens} & \textbf{\% Good} & \textbf{\% Bad} \\
\midrule
Original        & 2.91M & 67.2 & 32.8 \\
Random         & 2.91M & 49.8 & 50.2 \\
Balanced       & 1.40M & 49.5 & 50.5 \\
Bad-skewed     & 1.43M & 32.9 & 67.1 \\
Good-skewed    & 1.38M & 65.8 & 34.2 \\
Good-only      & 1.95M & 100.0 & 0.0 \\
Dense Balanced      & 1.82M & 50.0 & 50.0 \\
\bottomrule
\end{tabular}
\caption{
Token-level supervision statistics for label sensitivity settings. We report the total number of supervised tokens and the proportion of positive (factual) and negative (hallucinated) labels for each configuration.
}
\label{tab:label_stats}
\end{table}

\subsection{LoRA-A and LoRA-B Analysis for DPO and SOPL}
\label{appendix_DPO&SOPL}

As shown in Figure~\ref{fig:loraA_loraB_compare_DPO_SOPL}, we further analyze the LoRA-$A$ and LoRA-$B$ components learned by SOPL and DPO.

\begin{figure}[H]
\centering
% ===== Row 1: SOPL =====
\begin{subfigure}[t]{0.60\linewidth}
    \centering
    \includegraphics[width=\linewidth]{loraA_SOPL.pdf}
    \caption{SOPL (LoRA-A)}
\end{subfigure}
\hfill
\begin{subfigure}[t]{0.38\linewidth}
    \centering
    \includegraphics[width=\linewidth]{loraB_SOPL.pdf}
    \caption{SOPL (LoRA-B)}
\end{subfigure}
\vspace{0.5em}
% ===== Row 2: DPO =====
\begin{subfigure}[t]{0.60\linewidth}
    \centering
    \includegraphics[width=\linewidth]{loraA_DPO.pdf}
    \caption{DPO w/o SFT (LoRA-A)}
\end{subfigure}
\hfill
\begin{subfigure}[t]{0.38\linewidth}
    \centering
    \includegraphics[width=\linewidth]{loraB_DPO.pdf}
    \caption{DPO w/o SFT (LoRA-B)}
\end{subfigure}
\vspace{0.5em}
% ===== Row 3: SFT-first DPO =====
\begin{subfigure}[t]{0.60\linewidth}
    \centering
    \includegraphics[width=\linewidth]{loraA_SFT_DPO.pdf}
    \caption{DPO-after-SFT (LoRA-A)}
\end{subfigure}
\hfill
\begin{subfigure}[t]{0.38\linewidth}
    \centering
    \includegraphics[width=\linewidth]{loraB_SFT_DPO.pdf}
    \caption{DPO-after-SFT (LoRA-B)}
\end{subfigure}
\caption{
LoRA-$A$ and LoRA-$B$ analysis for DPO and SOPL.
% \textbf{Top row:} SOPL; \textbf{bottom row:} DPO.
% \textbf{Left column:} LoRA-$A$ separability heatmaps.
% \textbf{Right column:} LoRA-$B$ steering curves under different scaling factors $\lambda$.
% While both methods exhibit relatively weak LoRA-$A$ separability compared to TOPL, they still demonstrate clear LoRA-$B$ steering behavior.
\textbf{Top row:} SOPL; \textbf{middle row:} DPO trained from the instruction-tuned model; \textbf{bottom row:} DPO-after-SFT.
\textbf{Left column:} LoRA-$A$ separability heatmaps.
\textbf{Right column:} LoRA-$B$ steering curves under different scaling factors $\lambda$.
All three methods exhibit relatively weak LoRA-$A$ separability than TOPL. While SOPL and DPO trained directly from the instruction-tuned model exhibit clear LoRA-$B$ steering behavior, DPO-after-SFT shows weaker steering.
}

\label{fig:loraA_loraB_compare_DPO_SOPL}
\end{figure}

% Compared to TOPL, both DPO and SOPL exhibit substantially weaker LoRA-$A$ separability across layers and modules, suggesting that their learned low-rank subspaces provide less discriminative structure between factual and non-factual tokens. Interestingly, however, both methods still demonstrate clear LoRA-$B$ steering effects under scaling.

% This observation suggests that the LoRA-$B$ steering effect is not unique to TOPL; instead, the primary distinction appears to lie in the structure of LoRA-$A$. A possible explanation is that effective steering behavior depends not only on learning useful steering directions through LoRA-$B$, but also on learning appropriate conditions under which these directions should be activated. In particular, LoRA-$A$ appears to determine how selectively the learned steering directions are applied within the hidden representation space. Although DPO and SOPL learn usable LoRA-$B$ directions, their weaker LoRA-$A$ structure may limit how effectively these directions are utilized after merging. By contrast, the stronger LoRA-$A$ separability learned by TOPL may allow the model to more consistently activate the appropriate steering directions, leading to improved factuality and stronger OOD robustness.

Compared to TOPL, both DPO and SOPL exhibit substantially weaker LoRA-$A$ separability across layers and modules, suggesting that their learned low-rank subspaces provide less discriminative structure between factual and non-factual tokens. Interestingly, however, SOPL and DPO trained directly from the instruction-tuned model still exhibit clear LoRA-$B$ steering effects under scaling. In contrast, the steering effect becomes substantially weaker when DPO is applied after SFT. This difference suggests that the emergence of strong LoRA-$B$ steering under DPO can depend on the model initialization and optimization trajectory.

The strong LoRA-$B$ steering effect of SOPL and DPO trained from instruction-tuned model also suggests that the LoRA-$B$ steering effect is not unique to TOPL; instead, the primary distinction appears to lie in the structure of LoRA-$A$. A possible explanation is that effective steering behavior depends not only on learning useful steering directions through LoRA-$B$, but also on learning appropriate conditions under which these directions should be activated. In particular, LoRA-$A$ appears to determine how selectively the learned steering directions are applied within the hidden representation space. Although DPO and SOPL learn usable LoRA-$B$ directions, their weaker LoRA-$A$ structure may limit how effectively these directions are utilized after merging. By contrast, the stronger LoRA-$A$ separability learned by TOPL may allow the model to more consistently activate the appropriate steering directions, leading to improved factuality and stronger OOD robustness.

\subsection{Detailed Machine Translation Results}
\label{appendix_translation_results}

Table~\ref{tab:translation_results} reports the complete machine translation results on Qwen3-8B under both in-distribution (ID) and out-of-distribution (OOD) settings. Higher values indicate better translation quality.

\begin{table}[H]
\centering
\small
\begin{tabular}{lcc}
\toprule
Method & ID & OOD \\
\midrule
SFT & 0.8843 & 0.9192 \\
TLDR & 0.8860 & 0.9226 \\
Unlikelihood & 0.8606 & 0.8972 \\
TOPL (Ours) & \textbf{0.8947} & \textbf{0.9300} \\
\bottomrule
\end{tabular}
\caption{Detailed Machine translation results on Qwen3-8B. TOPL achieves the best performance in both ID and OOD settings.}
\label{tab:translation_results}
\end{table}
\end{document}